\documentclass{article}
\usepackage{graphicx} 
\usepackage{booktabs}
\usepackage{multirow}
\usepackage{microtype}
\usepackage{tabularx}
\usepackage{threeparttable}
\usepackage{amsmath}
\usepackage{cleveref}

\title{Research on Driving Scenario Technology Based on Multimodal Large Lauguage Model Optimization}
\author{Wang Mengjie, Zhu Huiping, Li Jian, Shi Wenxiu, Zhang Song*}
\date{May 2025}

\begin{document}
\begin{sloppypar}

\maketitle

\begin{abstract}
With the advancement of autonomous and assisted driving technologies, higher demands are placed on the ability to understand complex driving scenarios. Multimodal general large models have emerged as a solution for this challenge. However, applying these models in vertical domains involves difficulties such as data collection, model training, and deployment optimization. This paper proposes a comprehensive method for optimizing multimodal models in driving scenarios, including cone detection, traffic light recognition, speed limit recommendation, and intersection alerts. The method covers key aspects such as dynamic prompt optimization, dataset construction, model training, and deployment. Specifically, the dynamic prompt optimization adjusts the prompts based on the input image content to focus on objects affecting the ego vehicle, enhancing the model's task-specific focus and judgment capabilities. The dataset is constructed by combining real and synthetic data to create a high-quality and diverse multimodal training dataset, improving the model's generalization in complex driving environments. In model training, advanced techniques like knowledge distillation, dynamic fine-tuning, and quantization are integrated to reduce storage and computational costs while boosting performance. Experimental results show that this systematic optimization method not only significantly improves the model's accuracy in key tasks but also achieves efficient resource utilization, providing strong support for the practical application of driving scenario perception technologies.
\end{abstract}

\providecommand{\keywords}[1]
{
\textbf{\text{Keywords: }} #1
}


\section{Introduction}
With the rapid development of autonomous driving technology, driving scenario perception has become one of the core modules of intelligent driving systems\cite{yurtsever2020survey,chen2024end}. In practical applications, driving scenario perception needs to handle a variety of complex tasks, which often involve the fusion and analysis of multimodal data (such as images and text), placing extremely high demands on the model's accuracy, generalization ability, and real-time performance\cite{prakash2021multi}. However, although current mainstream large-scale multimodal models perform well, their outputs rely on specific prompt designs for particular scenarios\cite{zhao2024scene}. How to build high-quality datasets and enhance the model's generalization ability under extreme conditions remains an urgent issue. Additionally, its engineering application depends on efficient training and deployment methods.

To address these challenges, this paper takes specific multimodal scenario understanding tasks, such as "cone detection, traffic light status recognition, and speed limit recommendation," as examples and proposes a comprehensive optimization method for multimodal models. The goal is to improve the model's performance in driving scenario tasks while reducing computational costs to meet deployment requirements.

Firstly, in terms of prompt design, this paper proposes a dynamic prompt optimization strategy based on task scenarios and model inference results. This strategy dynamically adjusts prompts according to the content of the input images, enabling the model to focus more on task-critical information, thereby significantly improving the model's task adaptability and inference accuracy. For example, if the image contains "traffic lights" or "cones," the strategy selects relevant prompts from a predefined prompt library to provide targeted alerts for the current scenario. Compared with traditional fixed prompt methods, dynamic prompts are more flexible and targeted, especially in multitask scenarios.

Secondly, in terms of dataset construction, we adopt a hybrid approach using both real and synthetic datasets to address the high cost of data annotation and the long-tail distribution of driving scenarios. We leverage a VLM (Vision-Language Model) to mine and filter high-quality real multimodal data from a large-scale self-collected dataset. We then enrich the quantity and quality of the data using semi-supervised learning. Additionally, we use self-improving generative models and automated scene generation techniques to create diverse and realistic training datasets that include long-tail scenarios and objects, simulating different weather conditions, viewpoint changes, and obstacle insertions. After further supervised cleaning, these real and synthetic datasets are organized into high-quality datasets specifically for fine-tuning the model to enhance its performance and generalization ability.

Then, in terms of model training optimization, we propose a supervised hybrid method combining knowledge distillation and fine-tuning to efficiently improve model performance while reducing computational costs. The core of this method is to use a pre-trained large teacher model to transfer its knowledge to a smaller student model through knowledge distillation, enabling the student model to inherit the teacher model's strong performance. Meanwhile, we use LoRA technology to fine-tune key layers, further enhancing the student model's adaptability to specific tasks. This hybrid training approach not only improves model performance but also reduces overfitting risks. Experimental results show that this method performs well across multiple tasks, especially in resource-constrained environments, where its efficiency and practicality stand out. This hybrid training method provides an efficient and effective solution for optimizing large models, particularly suitable for scenarios requiring rapid deployment of lightweight models with limited task data.

Finally, to further reduce the storage and computational complexity of the optimized model, we employ advanced model quantization techniques to convert model weights and activations from floating-point to low-precision representations. This strategy is crucial for efficiently deploying deep learning models on resource-constrained devices, as it significantly speeds up model inference and improves overall system throughput. In this study, we specifically use AWQ (Activation-aware Weight Quantization). AWQ analyzes the distribution characteristics of activations in each layer to intelligently adjust the weight quantization strategy, effectively minimizing quantization errors and ensuring that the model maintains excellent performance even with reduced precision. Our experimental results show that the AWQ-quantized model meets resource constraints with almost no loss of original accuracy, demonstrating the method's great potential and value in practical applications. This approach not only improves computational efficiency but also provides a new pathway for the widespread application of large-scale deep learning models.

The optimization methods proposed in this paper address the shortcomings of existing technologies in dataset construction, model efficiency, and generalization ability, demonstrating the following innovative advantages: 1) Scene-aware synthetic datasets provide high-quality and diverse training data; 2) Dynamic prompt generation significantly enhances model task adaptability; 3) Hybrid optimization strategies for large models achieve a balance between high performance and low computational costs. Experimental results show that the proposed methods achieve significant performance improvements in tasks such as cone detection, traffic light status recognition, and speed limit recommendation, laying a solid foundation for the practical application of driving scenario perception technologies.

\section{Related Work}
In the field of driving scenario perception, the optimization and deployment of multimodal models have always been research hotspots. This section will review the existing research progress from three aspects: dataset construction (including data augmentation and synthetic data generation), prompt optimization, and model optimization, and compare the innovations of the methods proposed in this paper.

\subsection{Semi-Supervised Learning for Real Data Mining}
Semi-supervised learning leverages large amounts of unlabeled data to generate pseudo-labels, thereby enhancing model performance and showing significant potential in autonomous driving perception tasks. Early work, such as that by Hu et al.\cite{Hu_2022_CVPR}, utilized motion consistency constraints between consecutive frames and optical flow estimation to generate cross-frame pseudo-labels, improving the model's detection performance for occluded and blurred objects. Additionally, methods that use the geometric priors of LiDAR and 2D pseudo-labels from visual detection models through multi-view projection have been employed to enhance the quality of pseudo-label generation. Recently, with the development of Vision-Language Models (VLMs), VLMs have demonstrated potential in pseudo-label generation for autonomous driving. For example, VLM-MPC\cite{long2024vlm} reduces spatiotemporal drift in cross-modal pseudo-labels through temporal coherence constraints and spatial logic validation. ContextVLM\cite{sural2024contextvlm} generates zero-shot environment descriptions via prompt engineering and directly outputs structured labels. However, current research still faces challenges such as temporal drift of pseudo-labels in dynamic scenes, cross-modal semantic conflicts (e.g., "sunny" in vision versus "low light" in sensors), and the need for human expert intervention to verify the reliability of pseudo-labels in critical scenarios.

\subsection{Text-to-Image}
The development of Text-to-Image (T2I) models has gone through several stages, from early Generative Adversarial Networks\cite{goodfellow2020generative} to autoregressive models\cite{reddy2021dall}, and then to diffusion models\cite{ho2020denoising}, each bringing new breakthroughs to the T2I task. OpenAI's DALL-E model\cite{reddy2021dall}, based on the Transformer architecture, learns the joint distribution between text and images. These models have made significant progress in generation quality and controllability but still face challenges such as content safety and consistency. In recent years, diffusion models have become the mainstream technology in the T2I field, generating images through a gradual denoising process\cite{dhariwal2021diffusion}. Models like Imagen\cite{saharia2022photorealistic} and Stable Diffusion\cite{rombach2022high} use diffusion models to create images comparable to real photographs and human artworks. However, most of these methods primarily generate images from English text and lack sufficient support for Chinese. Kolors\cite{team2024kolors}, based on a diffusion model architecture, performs well in rendering both Chinese and English text, especially in high-resolution image generation. This paper fine-tunes and upgrades the open-source Kolors model to generate a diverse range of driving scenario images.

\subsection{Pure Rendering Data Synthesis}
Rendering engines are increasingly being used in autonomous driving data synthesis due to their realistic rendering effects and ease of scene construction. Traditional graphics engines, such as Carla\cite{dosovitskiy2017carla} and AirSim\cite{shah2018airsim}, rely on pre-built scenes or third-party modeling software. They excel in dynamics simulation but lack sufficient image realism and require SIL interfaces to reuse control algorithms. Neural rendering-based engines, such as MARS\cite{wu2023mars} and OISM\cite{wei2024editable}, achieve scene variations through data augmentation but are constrained by data quality and environmental lighting conditions. Image generation methods like BEVControl\cite{yang2023bevcontrol} support scene generation driven by control signals, but they suffer from poor spatiotemporal consistency and high training costs. Mainstream synthetic datasets, such as Synscapes\cite{wrenninge2018synscapes} and UrbanSyn\cite{gomez2023all}, face issues of either being closed (no details on generation) or having complex toolchains (requiring coordination of multiple software). The procedural data synthesis based on Blender proposed in this paper integrates all processes in a single tool, offers high-quality rendering, has a rich ecosystem, and can achieve scene diversity through parameterization.

\subsection{Prompt Engineering}
Prompt Engineering aims to guide Large Language Models (LLMs) to perform specific tasks more accurately by designing and optimizing prompts, without extensive adjustments to model parameters\cite{ekin2023prompt,sahoo2024systematic}. The field has evolved from initial zero-shot and few-shot prompting techniques to more complex multimodal knowledge distillation and automatic prompt generation. Pryzant et al.\cite{pryzant2023automatic} proposed a method combining gradient descent and beam search to systematically optimize prompts, further enhancing model performance . In this study, we iteratively optimize prompts based on specific task scenarios and model outputs, enabling general LLMs to achieve superior performance in specific tasks. This approach not only enhances the relevance and accuracy of model outputs but also expands the application scope and possibilities of Prompt Engineering.

\subsection{Model Distillation and Fine-tuning}
Model distillation and fine-tuning are common methods for enhancing model performance\cite{hsieh2023distilling,zhou2024empirical}. These techniques allow a well-trained, complex, and large-parameter teacher model to guide and optimize the learning of a simpler, smaller student model. Through fine-tuning, the student model can closely approximate the teacher model's performance, achieving state-of-the-art (SOTA) results with minimal human intervention and computational resources. For example, DistilBERT\cite{sanh2019distilbert}, a lightweight model distilled from BERT, retains similar performance while significantly reducing parameters and computational costs. Yang et al.\cite{yang2024survey} detailed the application of knowledge distillation in LLMs, including algorithm classifications, data augmentation techniques, and future research directions. QLoRA\cite{dettmers2023qlora}, which modifies only a small number of trainable parameters during fine-tuning while retaining most of the model's knowledge, achieves performance comparable to full fine-tuning. However, these methods mostly rely on highly generalized datasets, posing significant challenges for long-tail scenario optimization in autonomous driving. To address this, this paper proposes an innovative solution that combines synthetic data with a hybrid distillation and fine-tuning approach. This not only significantly reduces inference costs but also maintains high model performance, especially in terms of stability and reliability in complex and rare scenarios.

\subsection{Model Quantization}
Model quantization is a key technique for optimizing the efficiency of deep learning models, aiming to reduce computational resource demands and storage space. PTQ (Post Training Quantization)\cite{frantar2022gptq} directly quantizes weights after training, which is simple but may affect accuracy. QAT (Quantization Aware Training)\cite{chen2023overcoming} simulates quantization effects during training to mitigate performance loss due to quantization, thereby improving final model accuracy (Jain et al., 2021). AWQ (Activation-aware Weight Quantization)\cite{lin2024awq}, an emerging method, is particularly suitable for large models. It guides weight quantization by analyzing activation distributions to minimize quantization errors. Compared to traditional quantization strategies, AWQ not only considers the information of weights themselves but also incorporates changes in activations, making the quantization process more refined and accurate, and effectively maintaining model performance. Additionally, studies have shown that AWQ can significantly reduce model size and computational complexity while maintaining high accuracy, especially in complex natural language processing or computer vision tasks. This indicates that by choosing appropriate quantization strategies, large deep learning models can be efficiently deployed in resource-constrained environments.

\section{Method}
\begin{figure}
    \centering
    \includegraphics[width=1.0\linewidth]{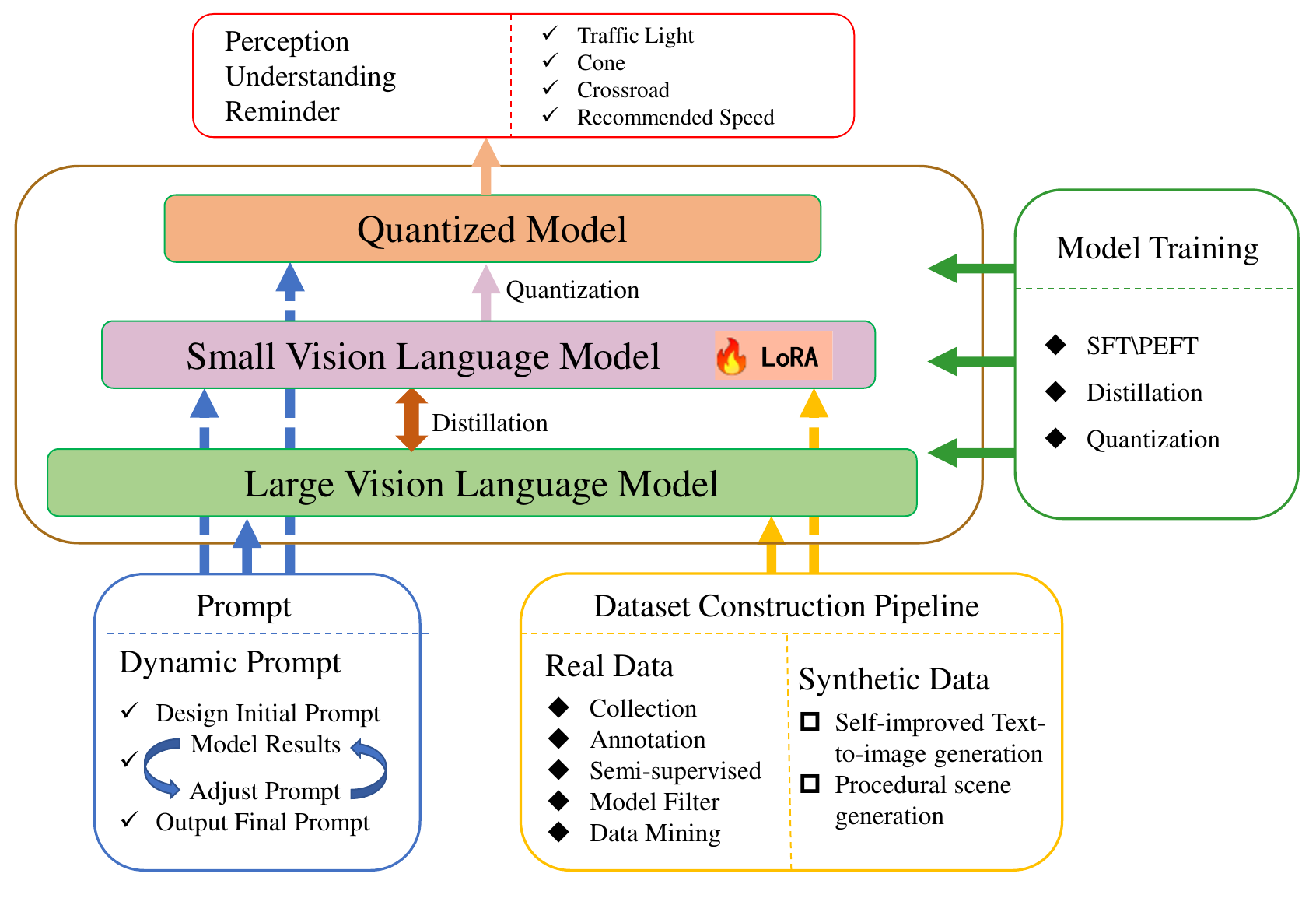}
    \caption{The overall architecture diagram proposed in this paper.}
    \label{fig:fig1-pipeline}
\end{figure}

The overall architecture of this paper is shown in \cref{fig:fig1-pipeline}.

\subsection{Dynamic Prompt Optimization}
The design of prompts directly affects the model's ability to understand tasks. First, it can significantly enhance the model's performance on specific tasks, transforming it from a "generalist with broad capabilities" to a "task specialist." Through carefully designed prompts, the model can better understand and generate task-related outputs without relying on large amounts of labeled data.

\begin{figure}[t]
    \centering
    \includegraphics[width=0.8\linewidth]{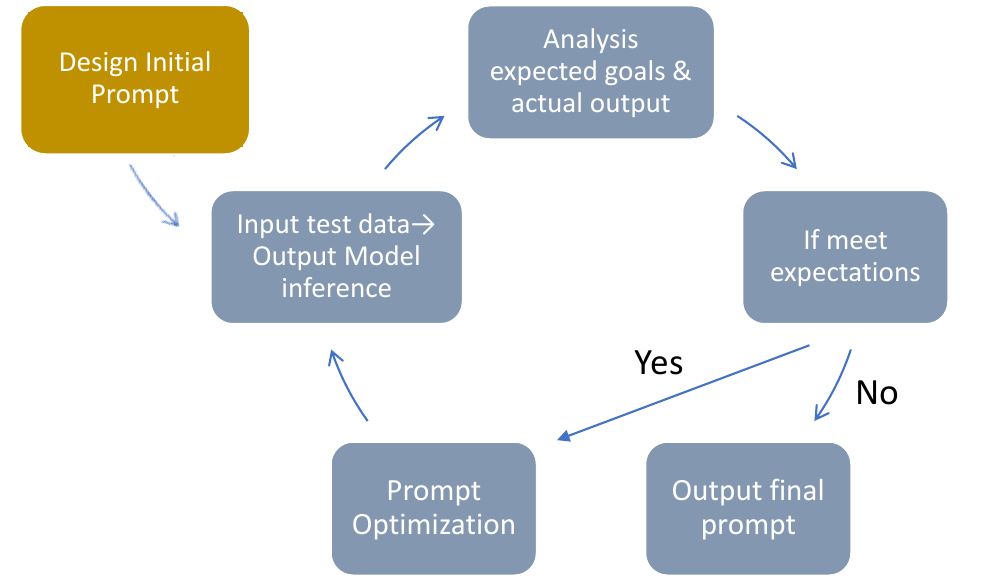}
    \caption{Dynamic Prompt Iterative Optimization Process.}
    \label{fig:fig2-prompt}
\end{figure}

In this paper, for tasks such as cone detection, traffic light status recognition, and speed limit recommendation, we adopt a dynamic prompt optimization strategy. This involves a systematic, multi-round iterative method to optimize model performance. We observe the model's outputs and adjust the prompts step by step until the desired effect is achieved. Initially, based on the preliminary prompts, we input a series of test data into the large language model and carefully observe the outputs. By comparing the differences between the expected goals and actual outputs, we make targeted adjustments to the prompts, such as clarifying task requirements, refining descriptions, or adding example guidance. Then, we conduct experiments with the updated prompts and repeat the process until the model's output meets the predetermined performance criteria. The specific process is shown in \cref{fig:fig2-prompt}. Specifically, for tasks that require precision, such as traffic light status recognition, we not only adjust the text descriptions but also introduce visual elements or multimodal information as auxiliary prompts to improve recognition accuracy. For the speed limit recommendation task, we focus more on the completeness and real-time requirements of contextual information to ensure that the generated prompts accurately reflect the speed limit rules in the current driving environment.

\begin{figure}[t]
    \centering
    \includegraphics[width=1.0\linewidth]{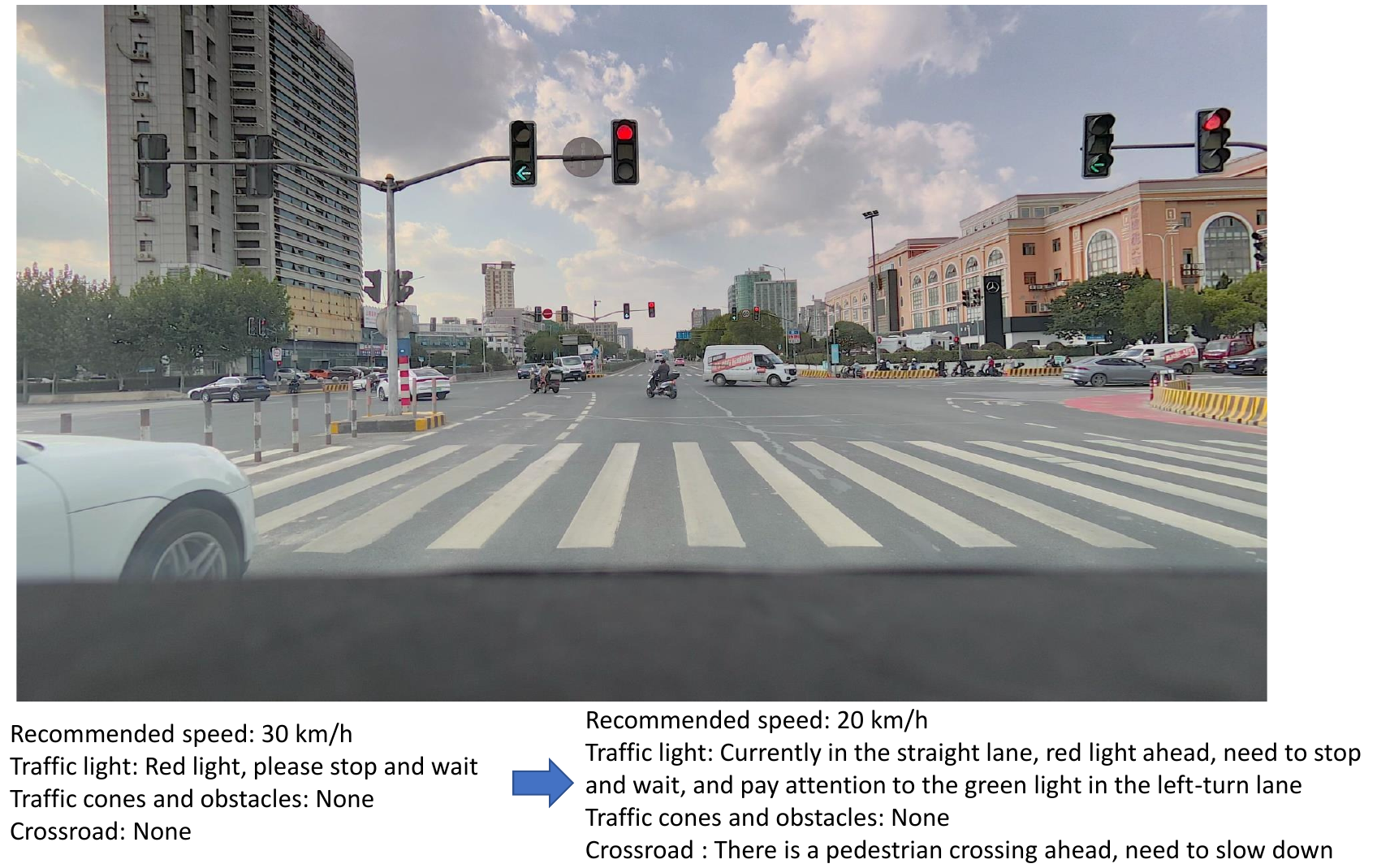}
    \caption{The model's output results before and after dynamic Prompt optimization.}
    \label{fig:fig3-prompt-result}
\end{figure}

Through this continuous iterative optimization method, we have achieved significant improvements in model performance across different application scenarios. \cref{fig:fig3-prompt-result} shows how dynamic prompt adjustment significantly enhances the performance of large models in specific tasks. In complex scenarios with multiple traffic lights at intersections, the model optimized with dynamic prompts can provide precise alerts based on the traffic light status of its own lane while also paying attention to the status of other lanes, offering drivers more comprehensive traffic information. Additionally, for zebra crossings at intersections, the optimized model can more accurately identify and remind drivers to be aware of pedestrians to ensure safe passage.

\subsection{Dataset Construction}
Dataset construction is the foundation of model training and directly affects the model's performance and generalization ability. To address the complex and diverse environmental conditions in driving scenario perception tasks, we propose a novel dataset construction method with the following core features: (1) Mining and labeling of real self-collected datasets; (2) Synthesizing scene-generalized datasets using text-to-image models and pure rendering techniques. We then filter and clean the datasets constructed by the above methods to obtain high-quality data to further enhance the model's performance and generalization ability.

\subsubsection{Self-Collected Real Data Mining}
Self-collected data typically covers the entire dataset, not just random samples. This comprehensiveness means that the data volume is much larger, providing richer and more detailed information. However, it also brings challenges, as self-collected data often contains various types of errors and outliers, increasing the complexity of the data. Therefore, when processing this type of data, extra caution is needed to ensure the accuracy and reliability of the data. To address these challenges, we propose an innovative data mining method using large models for analysis and labeling.

\begin{figure}
    \centering
    \includegraphics[width=1.0\linewidth]{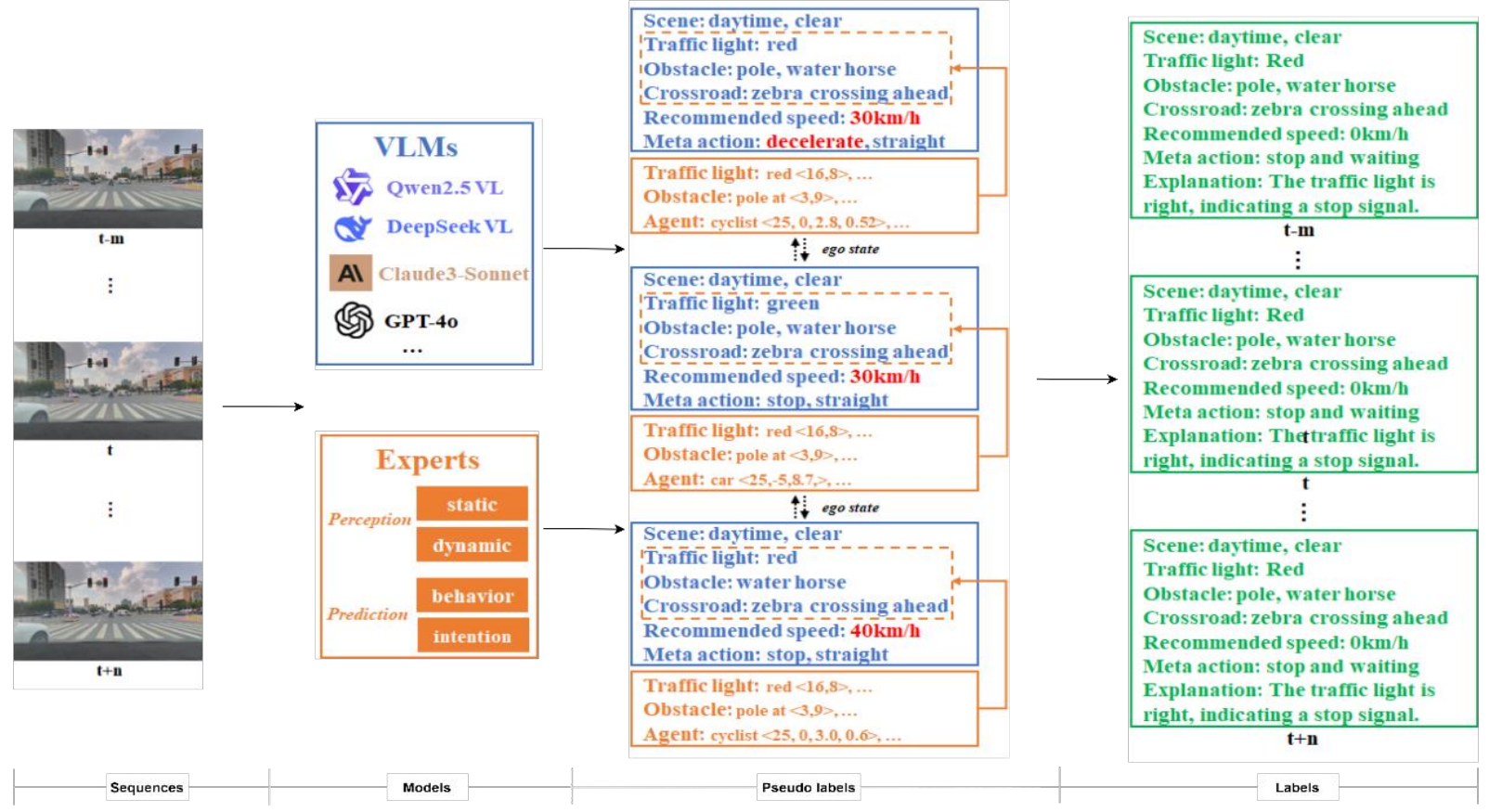}
    \caption{The flowchart of the self-Collected Real dataset construction process.}
    \label{fig:fig4-realdata-pipeline}
\end{figure}

This paper designs a multi-model joint pseudo-label generation framework with spatiotemporal constraints. By using temporal coherence constraints (label smoothness between adjacent frames) and spatial semantic consistency (e.g., matching traffic light status with vehicle behavior logic), we reduce pseudo-label noise and enhance the quality of pseudo-label generation through cross-validation between expert model and large model inference results. First, through prompt engineering, we set prompts and use the visual-language understanding and reasoning capabilities of Qwen2.5-VL to generate structured text labels for driving scenario tasks such as cone detection, traffic light status recognition, speed limit recommendation, and intersection alerts. Then, we perform collaborative filtering on the pseudo-labels across frames, applying temporal coherence and spatial consistency constraints. Specifically, we dynamically set a sliding time window based on vehicle speed (obtained from IMU). If the label of the current frame is inconsistent with the majority of labels in the time window, the current frame label is replaced with the majority label. Next, we correct obviously inconsistent pseudo-labels by checking whether the pseudo-labels for traffic lights, cones, and intersections match the vehicle's motion state. Finally, we use specialized perception models, such as traffic light and cone detection models, to generate expert pseudo-labels. By fusing the large model pseudo-labels with the expert pseudo-labels, we further improve the quality of the labels. The specific process is shown in \cref{fig:fig4-realdata-pipeline}.

\subsubsection{Data Synthesis}
Data synthesis can generate diverse scenarios, including different weather conditions, traffic situations, and geographical environments. This diversity helps cover long-tail scenarios in autonomous driving that are difficult to obtain, such as extreme weather or rare accident scenes. Synthetic data is highly controllable, allowing for editing and adjustment of elements in the scene, such as adding or removing objects and changing lighting conditions.

\begin{figure}
    \centering
    \includegraphics[width=1.0\linewidth]{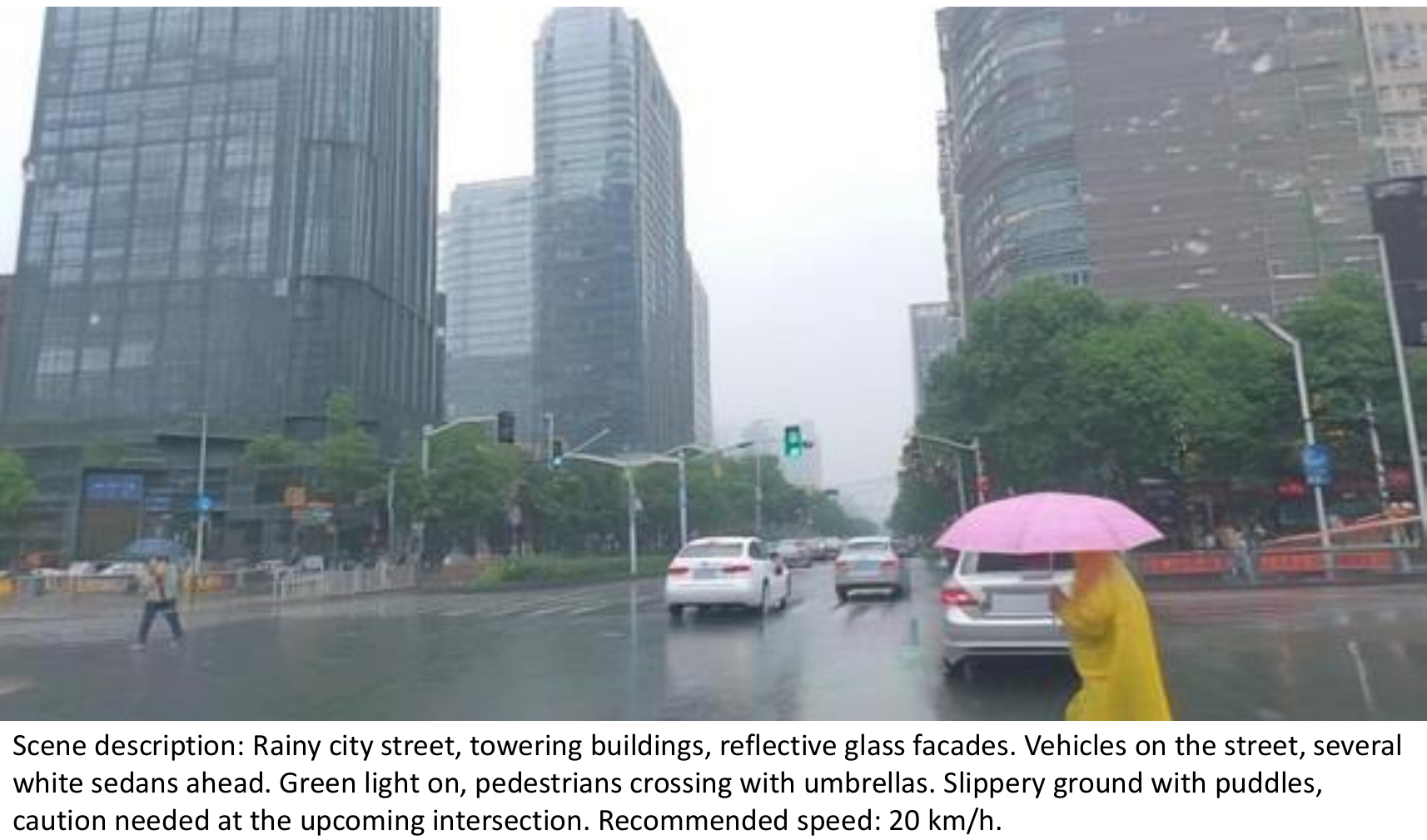}
    \caption{The images generated using text prompts by text-to-image methods.}
    \label{fig:fig5-t2i-result}
\end{figure}

\begin{figure}
    \centering
    \includegraphics[width=1.0\linewidth]{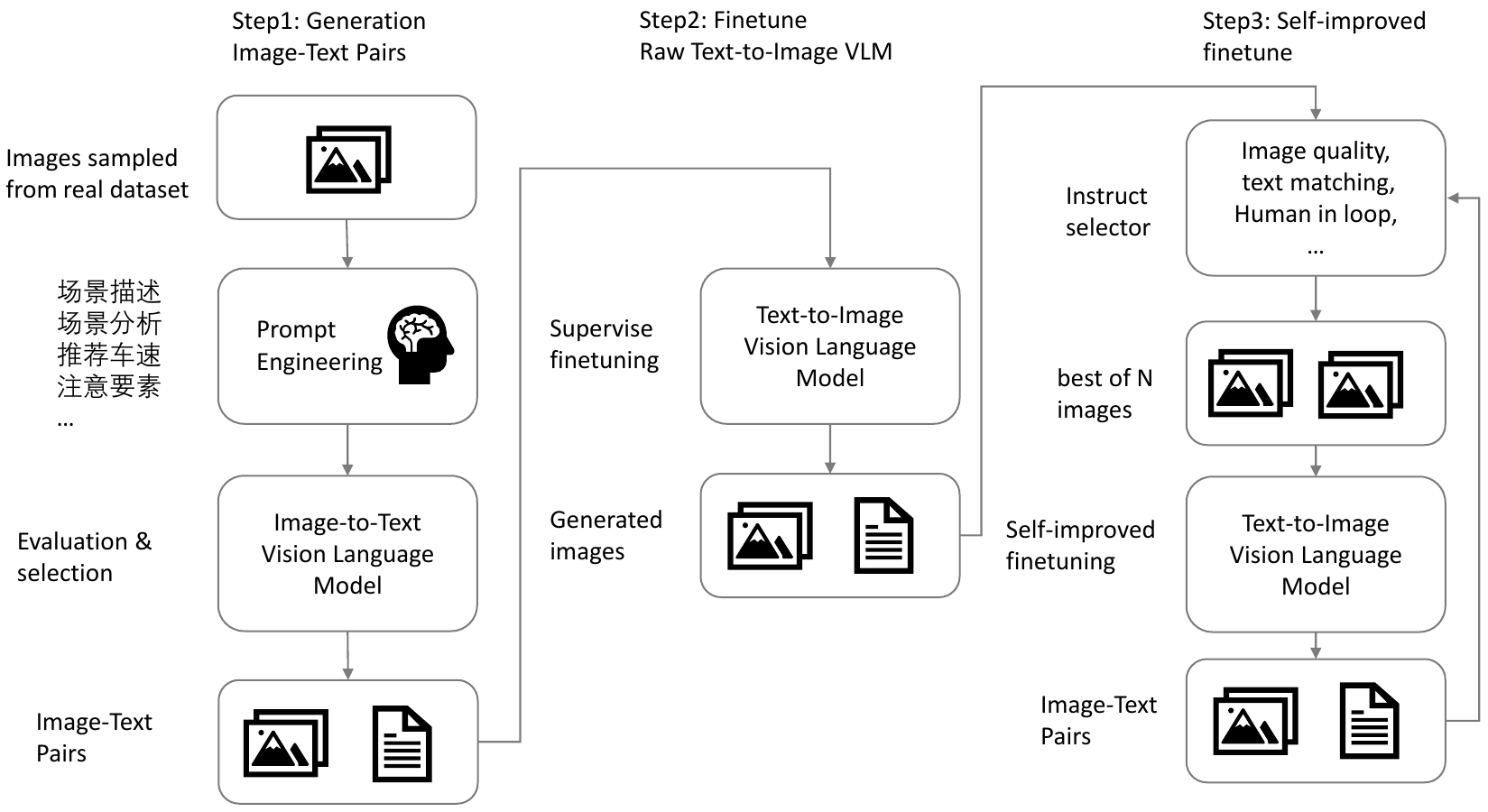}
    \caption{The text-to-image generation process employed in this paper.}
    \label{fig:fig6-t2i-pipeline}
\end{figure}

Text-to-Image:
In multimodal optimization tasks, Text-to-Image (T2I) technology meets the strict demands for data diversity and balance in autonomous driving scenarios by efficiently expanding special scenario datasets and alleviating sample imbalance. \cref{fig:fig5-t2i-result} shows an example generated by this technology: a pedestrian crossing the street with an umbrella in the rain. The self-improving image-to-text model training method proposed in this paper is shown in \cref{fig:fig6-t2i-pipeline}. First, initial training image-text pairs are generated. A batch of multimodal data containing special driving scenarios is systematically collected through a self-collected data mining system, including autonomous driving images and synchronized text descriptions. The text descriptions cover comprehensive scene descriptions (e.g., urban street views, buildings, weather, time), key elements (such as vehicle models, pedestrian dynamics, traffic signs, obstacles like cones, traffic light status), and road conditions (such as intersections, lane divisions). Then, the collected image-text pairs are strictly filtered to remove similar, repetitive, and misaligned data pairs, ensuring the data is representative and diverse, covering various autonomous driving scenarios and maintaining high quality. Next, an image-to-text large model automatically generates text descriptions for the image data to form an image-text pair dataset. Subsequently, the text-to-image model is preliminarily fine-tuned using LoRA (Lora-Rank Adaptation) technology. Finally, in the self-improving fine-tuning process, the preliminarily fine-tuned text-to-image model and text descriptions are used to generate multiple images for specific driving scenarios. The generated images are scored based on image quality and consistency with text descriptions, and the best N images are selected as new training data for fine-tuning the text-to-image model. These generated images greatly enrich the autonomous driving scenario dataset, effectively addressing challenging samples and sample imbalance issues.

\begin{figure}
    \centering
    \includegraphics[width=1.0\linewidth]{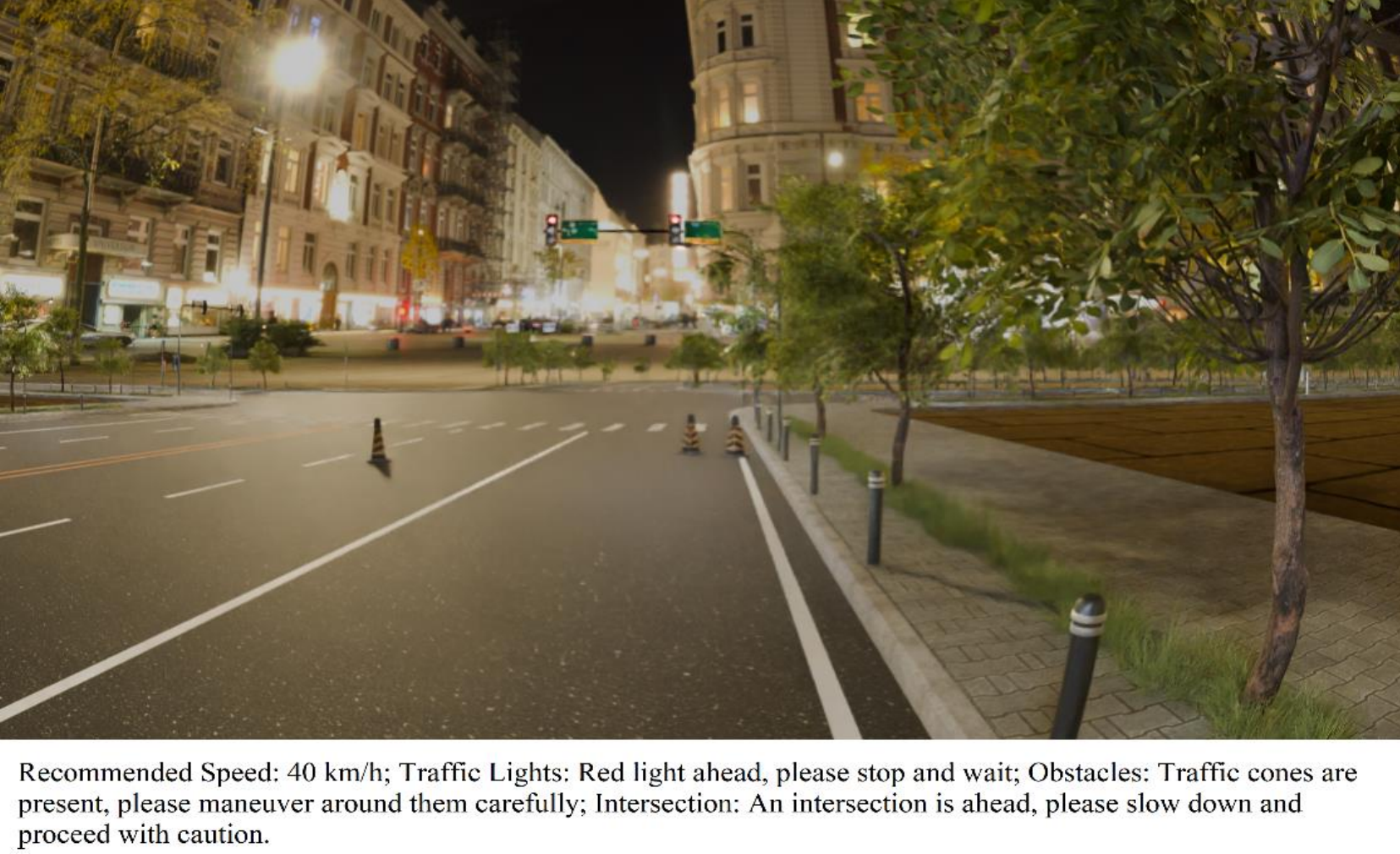}
    \caption{Images generated using pure rendering method.}
    \label{fig:fig7-blender-result}
\end{figure}

\begin{figure}
    \centering
    \includegraphics[width=1.0\linewidth]{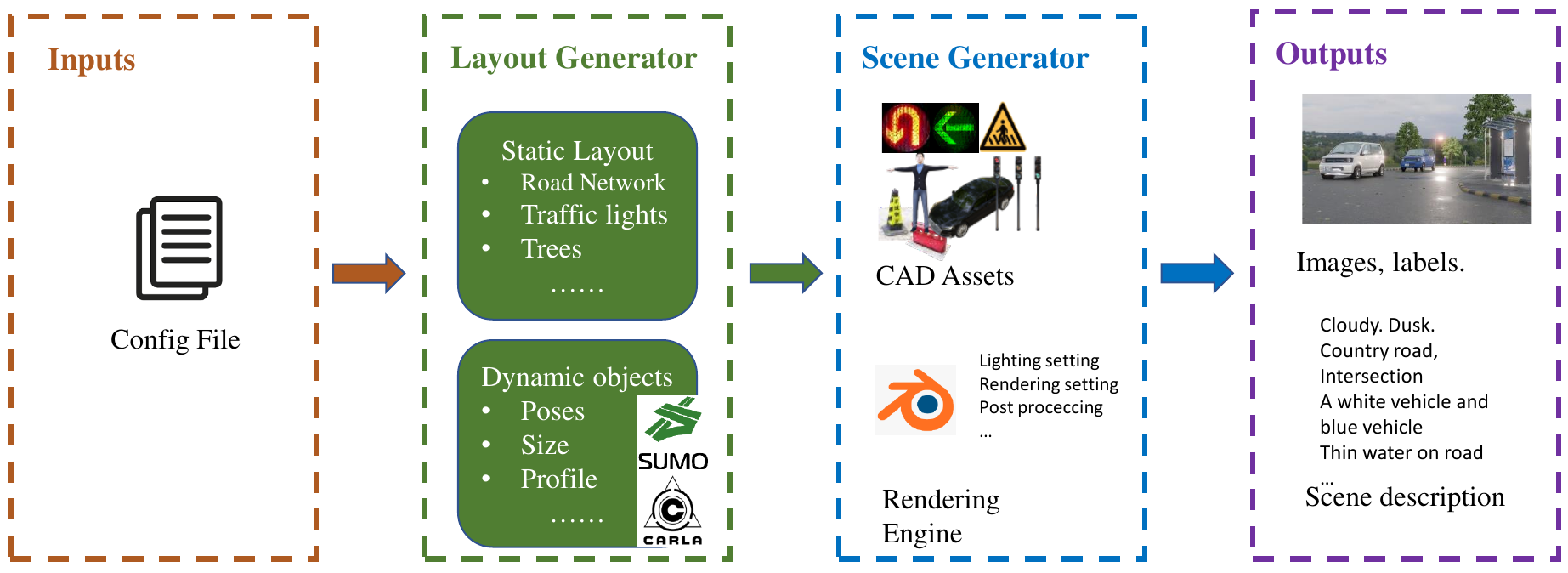}
    \caption{The data synthesis process using pure rendering methods.}
    \label{fig:fig8-blender-pipeline}
\end{figure}

Pure Rendering:
In the process of data synthesis using rendering engines, we leverage the Blender rendering engine to adjust scenes according to specific task requirements, designing highly task-matched scenarios to maximize the generation of positive sample data. For tasks such as cone detection and traffic light recognition, we can add more intersection samples during scene design, with results shown in \cref{fig:fig7-blender-result}. The key process of pure rendering data synthesis is shown in \cref{fig:fig8-blender-pipeline}. Initially, a structured static layout, inclusive of a road network, is generated based on configuration files. Simulators such as SUMO, CARLA, or LimSim, along with predefined rules, are employed to create trajectories or poses for dynamic objects. Given the task requirements, the ego vehicle is strategically positioned at intersections whenever possible, and obstacles like cones are randomly placed in front of the vehicle using CAD models. Subsequently, in the scene generation phase, a 3D scene is constructed in Blender by integrating the static layout with the dynamic trajectories and positional data, leveraging a 3D CAD asset library. Rendering and camera parameters are meticulously set within Blender to produce highly realistic image data, capitalizing on Blender's advanced rendering capabilities. Lastly, during the output label and prompt generation stage, detailed image scene descriptions and corresponding prompt information, tailored to our specific needs, are generated based on the constructed scene and ego vehicle positioning. For instance, for the sample image, the generated prompt reads: "Recommended Speed: 40 km/h. Traffic Lights: Red light ahead, please stop and wait. Obstacles: Traffic cones are present, please maneuver around them carefully. Intersection: An intersection is ahead, please slow down and proceed with caution."

\subsection{Model Distillation and Fine-tuning}
Model distillation and fine-tuning are crucial steps for optimizing the performance of multimodal models, especially when deploying high-performance models on resource-constrained edge devices. To achieve efficient and innovative goals, we employ a hybrid approach combining distillation and fine-tuning. By leveraging the knowledge of a large teacher model and optimizing with a combination of self-collected and synthetic data, we significantly enhance the student model's performance in complex driving scenario tasks. Specifically, we first use knowledge distillation to transfer the knowledge from the large teacher model to the student model. The teacher model is typically a high-performance but computationally complex model, while the student model is a less computationally demanding one. For our task scenarios, we select Qwen2.5-VL-72B-Instruct as the teacher model and Qwen2.5-VL-7B-Instruct as the student model. We use the teacher model to predict the dataset, obtaining the predicted probability distribution (soft labels) for each sample. We employ a hybrid loss function that combines soft label loss and hard label loss (both of which can be considered cross-entropy losses). The soft label loss encourages the student model to match the teacher model's output probability distribution, measured using KL divergence (Kullback-Leibler Divergence), as shown in \cref{eq:eq1}. The hard label loss encourages the student model to correctly predict the true labels. Subsequently, we adopt LoRA fine-tuning, which injects trainable low-rank decomposition matrices into the key layers of the Transformer to reduce the number of training parameters and approximate full-parameter fine-tuning. Based on the distilled model weights $\Phi_{0}$, we encode task-specific parameter increments $\Delta\Phi=\Delta\Phi(\Theta)$ with a smaller set of parameters $\Phi$, where $|\Theta|<<|\Phi_{0}|$. The task of updating $\Delta\Phi$ becomes the optimization of $\Phi$, i.e., maximizing the conditional model probability for parameter updates $\Theta_{0}+\Delta\Theta$, as shown in \cref{eq:eq2}. Finally, in the performance validation and optimization phase, we conduct a comprehensive evaluation of the hybrid model, covering metrics such as accuracy, recall, inference latency, and memory usage, to ensure its efficiency and adaptability in practical applications. This hybrid approach of distillation and fine-tuning is particularly suitable for scenarios that require rapid deployment of lightweight models with limited task data, especially in resource-constrained environments such as edge devices and mobile platforms, where it enables high-performance model deployment.


\begin{equation}
    D_{KL}(P||Q) = \sum_{i} P(i)log\frac{P(i)}{Q(i)} \label{eq:eq1}
\end{equation}

\begin{equation}
    \max_{\Theta} \sum_{(x,y)\in Z} \sum_{t=1}^{|y|}log(P_{\Phi_{0}+\Delta\Phi(\Theta)}(y_(t)|x,y<t)) \label{eq:eq2}
\end{equation}

\subsection{Model Quantization}
To further enhance computational efficiency, reduce storage requirements, and lower power consumption, we employ a quantization technique that converts high-precision parameters (e.g., 32-bit floating-point) to low-precision representations (e.g., 8-bit integers). This significantly reduces computational complexity and memory usage, enabling large-scale models to run efficiently on mobile devices or embedded systems. After in-depth analysis of various quantization methods and extensive ablation studies, we selected the most suitable approach for our needs. Specifically, we adopted the AWQ (Activation-aware Weight Quantization)\cite{lin2024awq} method, which adjusts the weight quantization strategy based on the distribution of activations to minimize quantization error. AWQ first analyzes the distribution characteristics of activations in each layer and selects the optimal quantization levels for each weight based on this information. This approach ensures that the model maintains high accuracy even with low-bit representations. Experimental results show that after applying AWQ to the optimized model, there is only a minimal performance drop compared to the full-precision model in tasks such as cone detection, traffic light alerts, and speed limit recommendations, while computational efficiency is significantly improved. Moreover, AWQ demonstrates better adaptability and accuracy than other quantization methods in handling complex tasks due to its unique activation-aware mechanism.

\section{Experiments}
To comprehensively evaluate the model's performance, we conducted extensive experiments on datasets that include both meticulously collected real data and synthesized data generated through text-to-image and pure rendering techniques. Initially, we trained the model using the full set of real data to explore its capabilities in practical application scenarios. Subsequently, to investigate the potential value of synthetic data in the training process, we further compared the model's performance when incorporating synthetic data into the training tasks versus using only synthetic data. Through this comparative analysis, we can more clearly understand the specific contributions of synthetic data to enhancing model performance.

\subsection{Dataset and Experimental Setup}
\cref{tab:tab1} shows the distribution of the training and validation sets for both self-collected and synthetic data. For real data, we used images captured by the ego vehicle and pre-annotated them with a 2D object detection model to select high-quality images relevant to the training tasks. For synthetic data, we employed two methods: first, using a text-to-image model to optimize scene-specific images based on optimized prompts; second, employing pure rendering techniques to generate the required multimodal datasets. We conducted experiments on eight NVIDIA A100 GPUs with a batch size of 10 to balance memory usage and training efficiency, ensuring efficient utilization of computational resources while maintaining the stability and accuracy of model training.

\begin{table}[htbp]
    \centering
    \caption{Data distribution}
    \begin{tabular}{ccc}
        \toprule
         & Train dataset & Test dataset \\
        \midrule
        Real data & 19360 & 2885 \\
        Text-to-image & 11486 & 2332 \\
        Pure Rendering & 11486 & 2332 \\
        \bottomrule
    \end{tabular}
    \label{tab:tab1}
\end{table}

\subsection{Main Results}
First, we validated the effectiveness of our proposed system architecture on real data. \cref{tab:tab2} shows that through prompt optimization, open-source model validation and deployment, model distillation and fine-tuning, and the application of synthetic data, the accuracy of cone detection, traffic light status alerts, and speed limit recommendations was significantly improved. Specifically, using the open-source 7B model and the quantized 7B model for inference, the average accuracy values were 0.580 and 0.542, respectively. Quantization led to a slight drop in model performance but a significant increase in inference speed, which benefits end-side deployment. With our proposed methods, without reducing the inference speed of the quantized model, the average accuracy was significantly improved, with an increase of 0.322 compared to the open-source quantized model. Additionally, based on the model trained on real data, we used synthetic data for fine-tuning to enhance the model's generalization ability by expanding data quality and quantity. Table 4 shows the model's performance on various tasks after fine-tuning with synthetic data. Overall, the average accuracy increased by about 0.022, with improved recognition capabilities across all tasks.

These experimental results demonstrate that our proposed large model optimization methods play a crucial role in accuracy and significantly enhance overall performance and reliability when executing specific tasks. Through optimization, the model can more accurately process complex input data and generate outputs that better meet task requirements. This not only improves task execution efficiency but also enhances the model's adaptability and stability in practical applications.

\begin{table}[htbp]
  \centering
  \scriptsize
  \caption{Performance of the model optimized using real data.}
    \begin{tabular}{cc|ccccc}
    \toprule
    \multicolumn{2}{c|}{} & 72B+Qua(official) & 7B    & 7B+Qua & our strategy$\star$ & our strategy \\
    \midrule
    \multicolumn{2}{c|}{FPS} & 20    & 8     & 9     & 11    & 11 \\
    \multicolumn{2}{c|}{Memory} & 89    & 60    & 87    & 87    & 87 \\
    \midrule
    \multirow{2}[2]{*}{Recommended Speed} & sMAPE & 0.092 & 0.186 & 0.226 & 0.101 & 0.086 \\
          & R2    & 0.818 & 0.578 & 0.41  & 0.763 & 0.809 \\
    \midrule
    \multirow{3}[2]{*}{Traffic Lights} & P     & 0.94  & 0.718 & 0.719 & 0.851 & 0.889 \\
          & R     & 0.943 & 0.688 & 0.683 & 0.885 & 0.903 \\
          & F1    & 0.941 & 0.703 & 0.701 & 0.868 & 0.896 \\
    \midrule
    \multirow{3}[2]{*}{Obstacles} & P     & 0.963 & 0.574 & 0.55  & 0.904 & 0.926 \\
          & R     & 0.992 & 0.624 & 0.584 & 0.904 & 0.914 \\
          & F1    & 0.977 & 0.598 & 0.566 & 0.904 & 0.92 \\
    \midrule
    \multirow{3}[2]{*}{Crossroad} & P     & 0.923 & 0.436 & 0.358 & 0.83  & 0.893 \\
          & R     & 0.906 & 0.439 & 0.361 & 0.809 & 0.84 \\
          & F1    & 0.914 & 0.437 & 0.359 & 0.819 & 0.866 \\
    \midrule
    \multicolumn{2}{c|}{Average} & 0.944 & 0.58  & 0.542 & 0.864 & 0.894 \\
    \bottomrule
    \end{tabular}%
    \begin{tablenotes}
      \scriptsize
      \item $\star$ indicates that the training set contains only self-collected real data.
    \end{tablenotes}%
  \label{tab:tab2}%
\end{table}%

\subsection{Visualization Analysis of Model Results After Fine-tuning with Synthetic Data}
\begin{figure}
    \centering
    \includegraphics[width=1.0\linewidth]{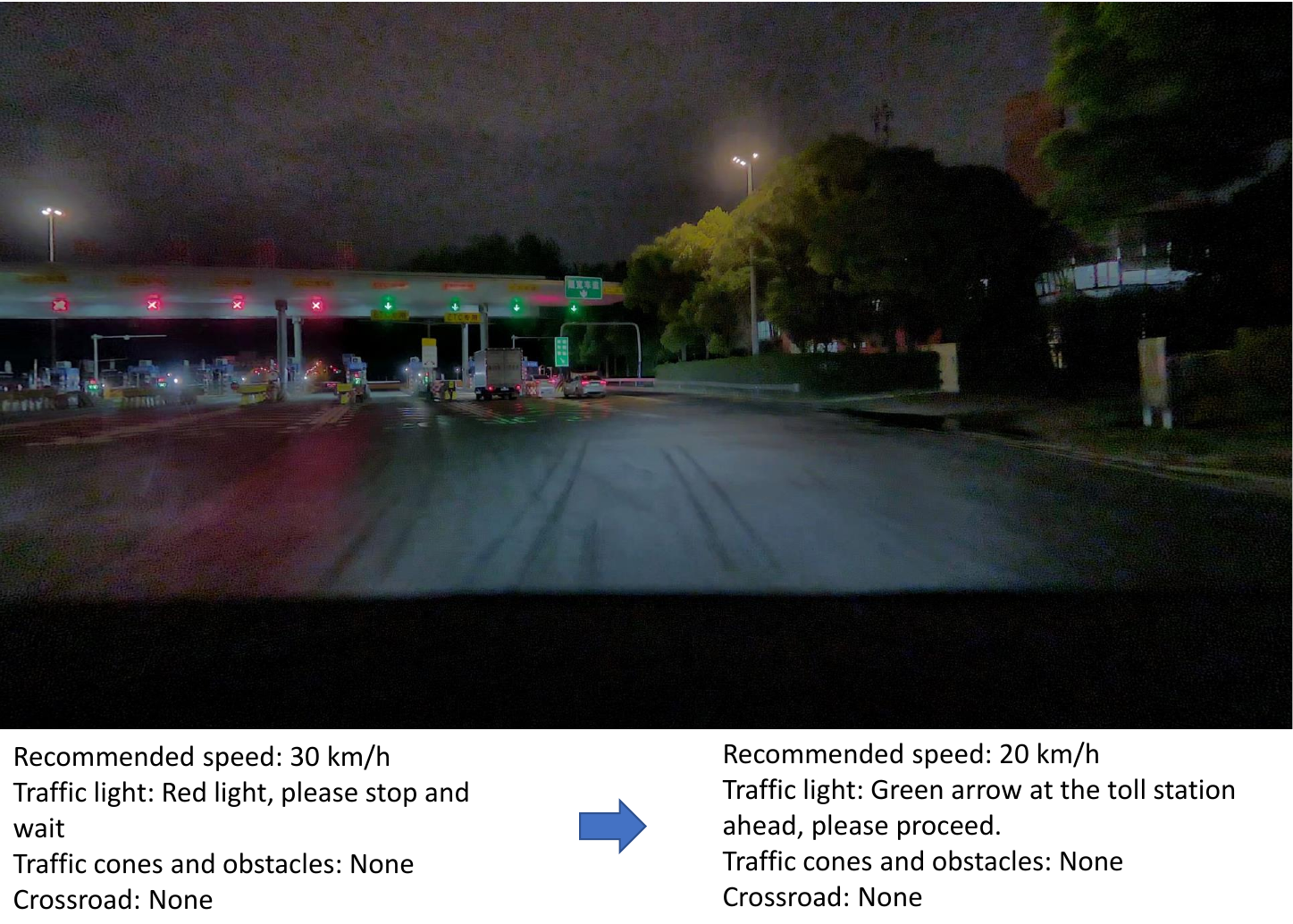}
    \caption{Long-tail scenarios of traffic lights that are easily confused.}
    \label{fig:fig9-visual}
\end{figure}

We further analyzed the impact of synthetic data on model performance. We tested long-tail scenarios that are prone to confusion. \cref{fig:fig9-visual} shows a traffic light scene that requires semantic understanding. The prompts below Figure 8 are the model outputs before and after fine-tuning with synthetic data. The results indicate that the model fine-tuned with synthetic data provides more accurate recognition of traffic lights. It can not only distinguish between traffic lights at intersections and those indicating whether a toll booth is open but also predict the traffic light corresponding to the future lane based on the current lane, thereby providing precise alerts.

\subsection{Ablation Studies}

\subsubsection{The Effect of Dynamic Prompt Optimization}
To verify the effectiveness of our proposed dynamic prompt optimization strategy, we used the official open-source Qwen2.5-VL-7B-Instruct model to generate corresponding text. \cref{tab:tab3} shows the experimental results before and after applying the dynamic prompt optimization strategy. It can be seen that after optimization, the model's output is more accurate, with an average precision increase of about 10\%. Specifically, for the tasks of speed limit recommendation and construction zone detection, which require strong scene understanding capabilities and consideration of current road conditions (including weather, traffic flow, road type, obstacles, etc.), the model's performance improved significantly when using the optimized prompts. The R2 score for speed limit recommendation increased by over 16\%, while the precision, recall, and F1 score for construction zone detection improved by 0.254, 0.171, and 0.222, respectively. For more common objects like traffic lights and intersections, which are relatively easier to recognize, the model's output also improved after prompt optimization, with F1 scores increasing by 0.013 and 0.059, respectively.

\begin{table}[htbp]
  \centering
  \scriptsize
  \caption{Model performance before and after prompt optimization.}
    \begin{tabular}{cc|cc}
    \toprule
    \multicolumn{2}{c|}{} & Original Prompt & After Prompt Optimization \\
    \midrule
    \multirow{2}[2]{*}{Recommended Speed} & sMAPE & 0.308 & 0.186 \\
          & R2    & 0.411 & 0.578 \\
    \midrule
    \multirow{3}[2]{*}{Traffic Lights} & P     & 0.687 & 0.718 \\
          & R     & 0.693 & 0.688 \\
          & F1    & 0.69  & 0.703 \\
    \midrule
    \multirow{3}[2]{*}{Obstacles} & P     & 0.321 & 0.574 \\
          & R     & 0.453 & 0.624 \\
          & F1    & 0.376 & 0.598 \\
    \midrule
    \multirow{3}[2]{*}{Crossroad} & P     & 0.366 & 0.436 \\
          & R     & 0.392 & 0.439 \\
          & F1    & 0.379 & 0.437 \\
    \midrule
    \multicolumn{2}{c|}{Average} & 0.484 & 0.58 \\
    \bottomrule
    \end{tabular}%
  \label{tab:tab3}%
\end{table}%

\subsubsection{The Impact of Different Synthetic Data Methods on Model Performance}
To evaluate the impact of data synthesized by different methods on model accuracy, we used two common approaches—text-to-image (T2I) and pure rendering—to generate synthetic data for a more generalized training dataset. We created an equal number of image-text pairs using each method. \cref{tab:tab4} compares the results of fine-tuning models trained on real data with datasets generated by each synthetic method. Overall, fine-tuning with either T2I or pure rendering data alone improved model accuracy, with average increases of 1.16\% and 2.23\%, respectively. However, when combining all synthetic data into the training set, the average accuracy improvement was more significant, reaching 3.2\%. Specifically, since pure rendering can place the ego vehicle at intersections as needed, the datasets for traffic lights and intersections are more extensive and of higher text quality. As a result, the model's detection F1 scores for these tasks were 1.64\% and 2.14\% higher, respectively, compared to fine-tuning with T2I data alone.

\begin{table}[htbp]
  \centering
  \scriptsize
  \caption{Impact of synthetic data on model performance.}
    \begin{tabular}{cc|cccc}
    \toprule
    \multicolumn{2}{c|}{} & our strategy$\star$ & Text-to-image & Pure Renderinf & our strategy \\
    \midrule
    \multirow{2}[2]{*}{Recommended Speed} & sMAPE & 0.101 & 0.098 & 0.1   & 0.086 \\
          & R2    & 0.763 & 0.773 & 0.77  & 0.809 \\
    \midrule
    \multirow{3}[2]{*}{Traffic Lights} & P     & 0.851 & 0.864 & 0.882 & 0.889 \\
          & R     & 0.885 & 0.889 & 0.904 & 0.903 \\
          & F1    & 0.868 & 0.877 & 0.893 & 0.896 \\
    \midrule
    \multirow{3}[2]{*}{Obstacles} & P     & 0.904 & 0.908 & 0.908 & 0.926 \\
          & R     & 0.904 & 0.911 & 0.909 & 0.914 \\
          & F1    & 0.904 & 0.91  & 0.909 & 0.92 \\
    \midrule
    \multirow{3}[2]{*}{Crossroad} & P     & 0.83  & 0.871 & 0.869 & 0.893 \\
          & R     & 0.809 & 0.81  & 0.835 & 0.84 \\
          & F1    & 0.819 & 0.84  & 0.861 & 0.866 \\
    \midrule
    \multicolumn{2}{c|}{Average} & 0.864 & 0.875 & 0.884 & 0.894 \\
    \bottomrule
    \end{tabular}%
    \begin{tablenotes}
      \scriptsize
      \item $\star$ indicates that the training set contains only self-collected real data.
    \end{tablenotes}%
  \label{tab:tab4}%
\end{table}%

\section{Conclusion}
This paper proposes a comprehensive system architecture for optimizing large models for specific tasks, significantly enhancing their performance and adaptability in the field of autonomous driving. The dynamic prompt optimization strategy flexibly adjusts prompts based on the content of input images, enabling the model to focus more on key information, thereby greatly improving task adaptability and inference accuracy. In terms of dataset construction, a scenario-aware enhancement scheme combining real and synthetic data effectively addresses long-tail distribution issues, enriches data diversity, and improves the model's understanding of complex scenarios. For model optimization, a hybrid method combining knowledge distillation and fine-tuning efficiently transfers knowledge from a large teacher model to a smaller student model, while LoRA technology enhances task-specific adaptability and reduces overfitting risks, particularly in resource-constrained environments. Finally, the AWQ quantization technique, which intelligently adjusts quantization strategies, effectively reduces model storage and computational complexity while maintaining high performance at low precision, providing strong support for the efficient deployment of autonomous driving technologies. The integrated application of these innovative methods offers new ideas and solutions for the development of autonomous driving.

\bibliographystyle{unsrt}
\bibliography{main.bib}

\end{sloppypar}
\end{document}